\documentclass[runningheads]{llncs}
\usepackage{graphicx}
\graphicspath{ {images/} }
\usepackage{amsmath,amssymb}
\usepackage{color}
\usepackage[width=122mm,left=12mm,paperwidth=146mm,height=193mm,top=12mm,paperheight=217mm]{geometry}
\usepackage{chngcntr}

\begin{document}

\pagestyle{headings}
\mainmatter

\title{High Diversity Attribute Guided Face Generation with GANs}

\titlerunning{High Diversity Attribute Guided Face Generation with GANs}

\authorrunning{High Diversity Attribute Guided Face Generation with GANs}

\author{Evgeny Izutov \\ \url{evgeny.izutov@intel.com}}
\institute{Intel \\ IOTG Computer Vision Group}

\maketitle

\begin{abstract}
In this work we focused on GAN-based solution for the attribute guided face synthesis. Previous works exploited GANs for generation of photo-realistic face images and did not pay attention to the question of diversity of the resulting images. The proposed solution in its turn introducing novel latent space of unit complex numbers is able to provide the diversity on the "birthday paradox" score 3 times higher than the size of the training dataset. It is important to emphasize that our result is shown on relatively small dataset (20k samples vs 200k) while preserving photo-realistic properties of generated faces on significantly higher resolution (128x128 in comparison to 32x32 of previous works).
\end{abstract}

\section{Introduction}

The last success of generative adversarial networks (GANs) in a wide range of utility tasks promotes researchers to pay more attention to GAN based solutions bringing each time something new. In this evidence the image generation task and the face synthesis problem in particular has become one of the most attractive challenges in computer vision. One must admit that other popular problems like classification, object detection, semantic segmentation have found their application in many real complex systems. As for image generation task, it lies between a non-standard test on the ability of modern neural networks to learn complex distributions on the one hand and some kind of art on the other hand.

\textbf{Motivation.} To overcome an unproductive situation with potentially powerful framework and to bring one more instrument into the machine learning development kit we believe researchers should focus their attention on the problem of diversity of generated images. It means that even if we have strong methods which are able to produce photo-realistic images we are still limited to select a couple of valid samples from the thousands of improbable ones for the real usage. Regarding the face synthesis and additional requirement to provide the diversity property over samples with shared spatial template it turns the task into the category of complex problems because of its insufficiency to copy the template of a face "two eyes, a nose and a mouth on the oval background" by combining different parts only. The real problem is that reasonable for people differences in faces are inseparable from the "mean face" for the most of generative models including GANs. In this evidence we should focus on developing the solution providing a good trade-off between the generative and discriminative properties of the objective function.

\textbf{Contribution.} To cope with the issue we propose to use the following feasible three-fold solution:
\begin{itemize}
\item Incorporating simple and powerful space of the unit complex numbers for the latent vector $z$ for GANs.
\item Extending well known histogram loss \cite{Ustinova:2016} to tackle a max entropy problem for the proposed below cycle-histogram.
\item Demonstrating the full model training procedure with auxiliary attribute classification subnetwork (inspired by \cite{Springenberg:2015} but improved in one critical way \--- overcoming possible gradient artifacts by using the regularization term from the \textit{contractive auto-encoder} \cite{Chen:2013}) and the identity preserving component which is built on the auto-encoder subnetwork with the auxiliary loss.
\end{itemize}

\section{Related Work}

\subsection{Face Synthesis}

Most of the existing methods exploit the  idea of fitting  parametric model of a face to the real one and then generate new face by changing key parameters. Depending on a parametrization such methods can be divided on 2 groups: methods with a direct parametrization which are trying to capture some task-specific property e.g. face shaving \cite{Nguyen:2008}, aligning \cite{Pengfei:2012} and face reconstruction \cite{Zhao:2014} and methods which can be used in performing transformations in a domain of face key points to carry out morphing between different facial expressions or a spatial rotation. Good survey for this can be found in \cite{LuZhihe:2017}.

The main limitation for the described above methods is a requirement to have a source face image. Another problem is to find a good model describing some properties like emotions, age, gender and so on.
Recent success of CNN-based solutions have shown significant improvements over the traditional CV methods in the domain of human faces (e.g. 3D face reconstruction \cite{Guo:2017}). Lately many researchers have focused their attention on the GAN-based solutions and it allows them to move problem from finding a good face model to learning a joint distribution of the required attributes. This fact makes the researcher be free from the model searching task and be more focused on the general solution.

\subsection{Generative Adversarial Network}

Introducing Generative Adversarial Networks (GANs) \cite{Goodfellow:2014} brought one more power framework to the family of generative models. Recent success of CNN based solutions has stimulated the usage of convolutional networks in the GAN framework but with some architectural changes like stride instead of pooling operators \cite{Radford:2015}, nearest neighbour upsampling versus deconvolutions \cite{Dong:2015}. It's important that, a GAN framework can be used not only for the direct image synthesis but also as an regularization loss \cite{Luc:2016}, \cite{Chen:2017}.

The GAN framework is based on a min-max game of two players: the first player (generator $G$) learns the distribution over the real data $p(x)$ with a domain $X$ by generating samples from the certain distribution over internal representation $p(z)$ with a domain $Z$ and the second one (discriminator $D$) tries to discriminate the generator answers from the real ones. More formally the GAN framework is presented in Eq. \ref{eq:1}.

\begin{equation}
\min_{\theta_G} \max_{\theta_D} E_{x \sim p(x)}[\log D(x)] + E_{z \sim p(z)}[\log(1 - D(G(z)))]
\label{eq:1}
\end{equation}

Usage of GANs is accompanied with difficulties in a training procedure due to its instability. To overcome them, two main ways are developed: some techniques applied during training like \textit{one sided label smoothing} \cite{Salimans:2016}, \textit{instance noise} \cite{Sonderby:2016}, \textit{gradient regularization} \cite{Roth:2017}, \textit{minibatch discrimination} \cite{Salimans:2016b}, \textit{distribution matching} \cite{Dumoulin:2016} and finding a cost function with better theoretical properties. Regarding the last one, the most impressive results were shown after rethinking the Wasserstein distance: WGAN \cite{Arjovsky:2017}, \cite{Gulrajani:2017} and with some simplification to the metric formulation BEGAN \cite{Berthelot:2017}. Other approaches use different discriminator scores \cite{Grewal:2017} or metrics \cite{Mroueh:2017}. The investigation of other cost functions and some generalized formulation of any function can be found in \cite{Nowozin:2016}.

The next barrier to successfully utilize GANs for real use-cases is a trade-off between quality and diversity. Several methods have been proposed to enhance an image quality by using some good working classical techniques like auto-encoders \cite{Larsen:2015}, \cite{Ulyanov:2017}, stacking lightweight and easy trainable subnetworks \cite{Huang:2016} and presented in a recent time the work of  \cite{Karras:2017} which proposes to grow generator/discriminator subnetwork to improve the situation with the gradient flow for the deep networks. Unfortunately, in practice the closer the output of an algorithm to the photo-realistic image the further from the diverse solution. To measure the variation of a generated data several approaches have been suggested: the inception score \cite{Salimans:2016} for the general purpose generators, explicit tests \cite{Metz:2016} and the birthday paradox test \cite{Arora:2017}. The last one looks more robust regarding its application for the wide range of target domains and it has been used in our work too. In general, variation decreasing is related to the "mode collapse" problem, which is proposed to solve by introducing the auxiliary network \cite{Srivastava:2017}, changing an objective function \cite{Metz:2016} or using some specific techniques \cite{Salimans:2016b}. In our opinion, such methods allow increasing the diversity but with too high cost  - restricting the network architecture or the training procedure. In this work we propose more simple yet effective solution which does not restrict general network architecture, meaning that last successful trends in network-building can be used here as well.

\subsection{Conditional Image Generation}

The next challenge beyond the arbitrary image generation task is a synthesis of images with specified attributes. The first GAN based attempts  to this task have been proposed in \cite{Mirza:2014}, \cite{Springenberg:2015} and its application for the face generation in \cite{Gauthier:2014}. Extended variant of the conditional-GAN framework is presented in Eq \ref{eq:2}, where $y$ is drawn from the distribution over the attributes with domain $Y$:
\begin{equation}
\min_{\theta_G} \max_{\theta_D} E_{x,y \sim p(x,y)}[\log D(x \mid y)] + E_{z \sim p(z), y \sim p(y)}[\log(1 - D(G(z \mid y)))]
\label{eq:2}
\end{equation}

Example of using the dense attributes like a semantic layout have been shown in \cite{Karacan:2016}. To control the attributes directly the auxiliary classification subnetwork is added \cite{Lee:2017}. However in the specified above approaches the main limitation for generation of valid samples for the each state of latent vector $z$ is an inability to generate samples from the learned distribution $\tilde{p}(z)$. It is because the distributions are different but we know the theoretical $p(z)$ only. To solve this problem it's been proposed to complement the network with auto-encoder like in \cite{Perarnau:2016}, add an auxiliary GAN to control the distribution of $z$ \cite{Zhang:2017} or to control the latent vector subspace by introducing FaceNet-based \cite{Schroff:2015} the identity preserving optimization \cite{Antipov:2017}, decomposing a latent vector into the internal $z_a$ and the attributes guided vector $y$. One more solution is to specialize some latent variables by the mutual information maximization \cite{Chen:2016}.

The different attempt to solve the attribute changing problem is based on training a specialized generator which is able to carry out only one type of a transformation. For instance, above mentioned model can translate any face image to some specified by the set of attributes face like  "man, black hair, square glasses" \cite{Wang:2017}, \cite{Kim:2017}, \cite{Zhu:2017}, \cite{Dong:2017}. For other set of attributes one more generator should be trained. The advanced methods allow researchers to carry out training on the unpaired data like in \cite{LuYongyi:2017} or \cite{Zhou:2017b}.

In the presented work we trying to improve the situation with a coherence between the learned $\tilde{p}(z)$ and real $p(z)$ distributions.


\section{Proposed Method}

\begin{figure}
\centering
\includegraphics[width=.3\textwidth]{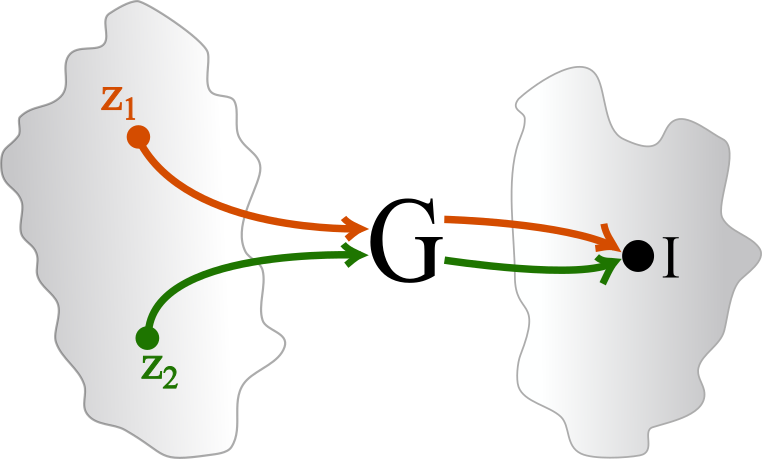}
\hspace{40pt}
\includegraphics[width=.3\textwidth]{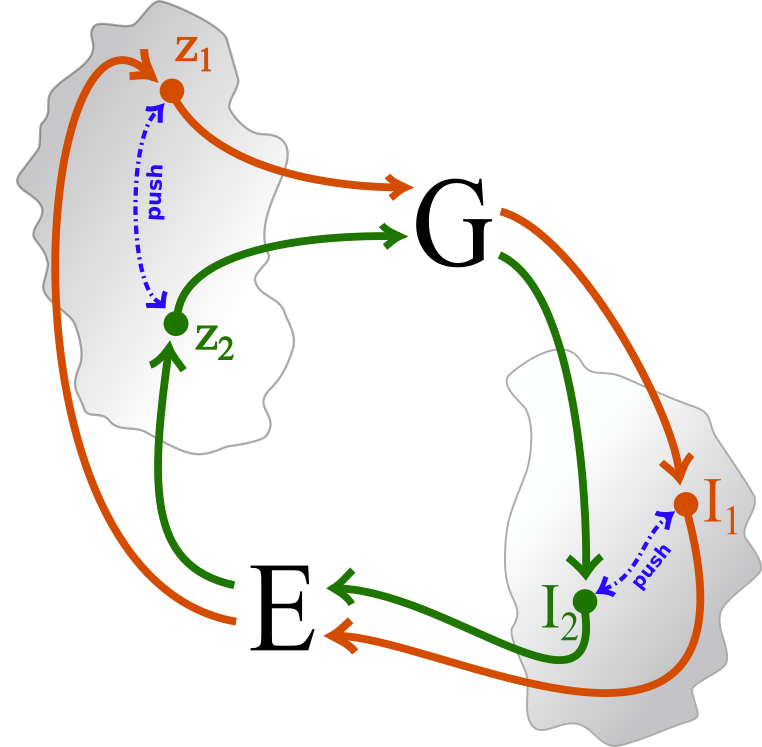}
\caption{Left: example of a mode collapse problem. Right: introducing auxiliary function $E$ incorporates an additional force to push different points far from each other.}
\label{fig:problem}
\end{figure}

\subsection{Background}
One of the problems with the original formulation of GAN framework (see Eq.\ref{eq:1}) is a \textit{mode collapse} (the left image on Fig. \ref{fig:problem}) \cite{Goodfellow:2017} which occurs when generator $G$ instead of learning bijective function $G: Z \mapsto I$ where $I$ is a space of generated images starts to map the different points from the latent vector space $Z$ to the same output point. This problem is aggravated by over-fitting while optimizing to get more photo-realistic images.

The reasonable assumption is that the choice of an objective function influences the properties of the obtained solution. Recently, it was proposed to use functions with more desirable properties like Wasserstein \cite{Arjovsky:2017}, \cite{Berthelot:2017} distance or more general f\--GAN \cite{Nowozin:2016} which is based on general $\mathcal{F}\--divergence$.

However, the last progress in theoretical analysis of GANs for the finite sample sizes \cite{Arora:2017b} showed some justification that low-capacity discriminator is not able to distinguish the target distribution from a learned distribution with high diversity. In other words even if the neural net distance which is modeled by the discriminator $D$ between distributions is small enough the real distribution $X$ is still can be far from the learned $I$. It's worth saying that it's not only a problem of an objective function selection but it's mainly a problem of a general formulation of the task which is defined for the finite capacity models and given  finite number of samples.

The key strategy is to take auxiliary an objective function for control the diversity properties directly instead of digging for more pleasant distance metric.

In our opinion the most suitable framework for this is an auto-encoder (see the right image on Fig. \ref{fig:problem}). It introduces an additional function $E: X \mapsto Z$ and the reconstruction distance between real sample $x$ and recovered from the internal representation state $G(E(x))$:
\begin{equation} \label{eq:3}
\min_{\theta_E, \theta_G} E_{x,y \sim p(x,y)}[l(x, G(E(x) \mid y))],
\end{equation}
where $l: X \times I \mapsto R$ is a loss function like $L_1$ or $L_2$ distances in common ways.

In context of GANs this loss is responded for the identity preserving objective: extracting and saving the most representative features during the main GAN based reconstruction phase. Incorporating an additional loss from Eq. \ref{eq:3} to our target objective (Eq. \ref{eq:2}) doesn't solve neither the problem of a coherence between the estimated and real distributions over the internal representation $z$ nor the diversity challenge. As it will be shown later in this chapter some additional loss which is connected with encoder $E$ can bring us the desired properties.

\subsection{Problem definition}

\begin{figure}[t]
\centering
\includegraphics[width=.3\textwidth]{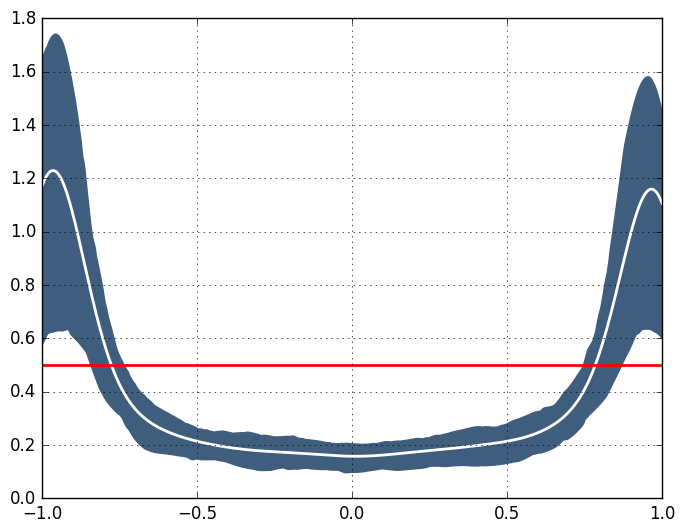}
\hspace{20pt}
\includegraphics[width=.3\textwidth]{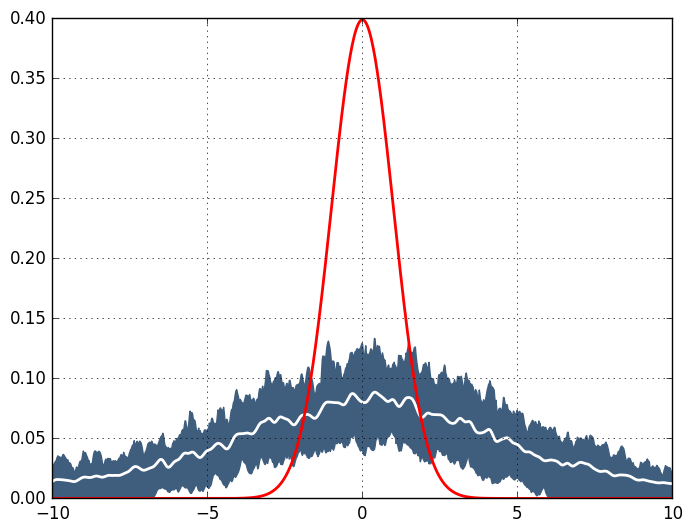}
\caption{Densities of the latent space variables in case of common used uniform and normal distributions. The red color curve is the theoretical distribution and white with blue bars - learned distribution. Left: uniform distribution with support $[-1; 1]$. Right: normal distribution $N(0; 1)$. As it can be seen there is a significant gap between theoretical and real distributions of latent vector $z$.}
\label{fig:src_z_dist}
\end{figure}

Bellow we assume that the domain of a distribution over the internal representation or \textit{latent} state $p(z)$ is $Z=R^d$. When we speak about auto-encoders the first question for us is the choice of the target space $Z$ because it implicitly influences on the capacity of a target model unlike more clear for understanding the size $d$ of a vector $z$. The most common practices are to choose $p(z)$ either from the uniform $U[-1, 1]$ or normal $N(0;1)$ multidimensional random variables. VAE \cite{Kingma:2013} uses slightly different formulation which estimates parameters of the selected distribution directly like $\mu$ and $\sigma$ in case of the normally distributed random variable.

The main limitation for usage auto-encoders together with GANs is a significant discrepancy between the learned distribution and the target distribution $p(z)$. It means that a generator $G$ is faced with learning two different distributions: one is output of the encoder $E$ (see Eq. \ref{eq:3}) and another one is GAN specific from Eq. \ref{eq:2}. In this case we can assume that two targets of objective function proposed above are independent. Moreover, in light of the limited capacity of a generator, the final model will be a trade-off between two losses and potentially will produce a poor performance compared with the single GAN objective.

One more problem is a degradation of the distribution learned by encoder. Fig. \ref{fig:src_z_dist} shows the theoretical and real distributions of $z$. For instance, the uniform continuous distribution with support $[-1, 1]$ is reduced into the uniform discrete distribution with only two different states $\{-1, 1\}$. Note that, neither the proper initialization of an encoder weights nor incorporating in a training process an additional loss to fit the target distribution (like auxiliary GAN for the latent space \cite{Zhang:2017} or as it's proposed later in this paper) can help to fix it.

In literature, the possible way to tackle this problem is to normalize the vector $z$ to the unit sphere. However, to preserve the maximal capacity property the individual variables in a vector $p(z)$ (which is a multivariate random variable) should be independent but after the normalization the condition mentioned above is broken.

Our goal is to select a space of a latent vector $Z$ for which we will be able to learn the same distribution of an encoder as the input distribution $p(z)$ for the generator during the GAN objective optimization.

\subsection{Proposed latent space}

In this paper regarding specified above problem about $Z$ space selection we focus on the uniform continuous distribution with a domain $[-1, 1]$. It has several advantages over the normal distribution or VAE \cite{Kingma:2013} technique: a linear dependence between the edge values $\textit{on} \equiv 1$ and $\textit{off} \equiv 1$ which can help during the optimization procedure and simple form of the support in context of the next synthesis task.

However, using it directly we face a problem of the saturation of the edge values: instead of learning the continuous manifold which is able to represent quantitative properties for each target feature the model learns qualitative properties only. Mostly it's result of the learning procedure which first extracts the most simple features and then concatenates more specific features or tries to clarify existing one. Unfortunately during the first stage the learning procedure is stumped near the edge values and we explored only one solution for this \--- to increase the size $d$ of the latent vector.

We do not consider the option of improving learning procedure by switching to adaptive methods or utilizing some other tricks because it’s a heavy solution without a guarantee of success. More clear way is to think not about the \textit{line segment} but about the \textit{closed segment}. Imagine if we don't have the edge values we can't stuck in them and all values can be covered by the forward step (a walk with positive move) only.

While being intuitive an idea to use the periodic functions like trigonometric $\sin(z)$ to achieve the desirable properties we should take into account that without the \textit{reference point} on the support, the feature cannot be used as a marker of some property and there is no continuous transformation between different states of the same feature. As a result, the final manifold will not satisfy the natural requirements of learning procedure which is SGD in our case.

In this work we propose to use the vector of the unit complex numbers as a domain of $Z$:
\begin{equation}
Z = \{z \in C^{d}: z = (x,y), x \in R, y \in R, x^2+y^2=1\}
\label{eq:4}
\end{equation}

Despite doubling the size of a latent vector $z$ to $2d$ the support size is still $d$ and can be interpreted as an angle $\alpha$ of a rotation of a unit vector. Continue our reasoning about the uniform distribution over the linear segment we have extended it to an angle formulation only. As we can see the proposed formulation saves the property of preserving the \textit{reference point} \--- we still are able to saturate at the edge values of a pair $(x,y)$ when $x$ or $y$ is close to the infinity and it corresponds to four angles $0, \pm \frac{\pi}{2}, \pi$ but thanks to the normalization term, so now we can force the state to leave the edge value.

Regarding its application for GAN this is expressed in the following:
\begin{itemize}
\item During sampling $z$ from $p(z)$ we first sample a random angle $\alpha \in [-\pi, \pi]$ and then project it on two axes to get $(x,y)$ pair.
\item To fit the output of an encoder $E$ to the introduced space $Z$ we double the output vector $z$ and then normalize each two values into the unit sphere.
\end{itemize}

\subsection{Additional losses}

As it was discussed above it's not enough to add the encoder part to the framework without additional losses which gives us two important properties: the uniform distribution over the angles in the latent space $Z$ and the diversity of the generated by $G$ samples in order to overcome the \textit{mode collapse} problem.

To tackle the \textbf{first problem} we borrowed the idea to estimate the $k$-dimensional histogram $H$ over some statistics from \cite{Ustinova:2016} and adopted it to the cycle distribution of angles $A=\{\alpha_1, ..., \alpha_N\}$. As in the original paper we assume the histogram nodes  $t_1=-\pi, t_2, ..., t_k=+\pi$ are uniformly spread out over the interval $[-\pi, +\pi]$. Then the value of each node $h_r$ is calculated in the same way but $t_{0}=t_{k}$ and $t_{k+1}=t_{0}$ due to the cycle property. After this we can interpret the histogram of angles as a discrete distribution and demand the uniform distribution over it. It can be simply reformulated as maximizing the entropy for the distribution of angles due to its property.

To solve the \textbf{second problem} we follow the proposed in \cite{Ulyanov:2017} idea and use the introduced earlier encoder $E$ with its acquired properties. The original concept was to use an encoder as an auxiliary component to align the latent space distributions of the real and generated data. We have extended it to control the diversity of the generated by $G$ samples by minimizing the reconstruction loss $l(;)$ between the real and estimated latent vector $z$. The key moment is to carry out the optimization in respect to the parameters of a generator only. It's very important to fit the parameters of an encoder on the real samples only. In other words an encoder should never see any sample from the space of synthesized points $I$. See in Eq. \ref{eq:8} formal description of the proposed loss.
\begin{equation}
\min_{\theta_G} {} E_{z \sim p(z), y \sim p(y)}[l(z, E(G(z \mid y)))] + E_{x,y \sim p(x,y)}[l(E(x), E(G(E(x) \mid y)))]
\label{eq:8}
\end{equation}
The first term is responded for the identity preserving in respect to the random latent state. The second one carry out the same thing but for the real sample and its representation state $z$ in a space $X$. Both terms complement each other to solve the above specified problem to combine in one model the distributions produced by encoder and real $p(z)$.

Finally, the objective function for the default GANs is a sum of terms of Eq. \ref{eq:2}, Eq. \ref{eq:3}, Eq. \ref{eq:8}. Note, in this paper we use BEGAN framework and that means that Eq. \ref{eq:2} is slightly different (for more details see the next chapter).


\section{Implementation details}

\subsection{Network architecture}

The proposed network consists of four main components: encoder $E$, generator $G$, discriminator $D$ and attribute classifier $A$ (see on Fig. \ref{fig:architecture}). As it was described earlier in \cite{Dong:2015} the discriminator subnetwork should require some additional conditions because the generator subnetworks is learned on the gradients back-propagated from the upper networks(which is the discriminator originally). It means that underlying subnetworks can learn all possible gradient artifacts, whose main sources are the checkerboard-based operations like max-pooling, deconvolution and convolution with stride more than one.

In other words if we want to use any network as objective $F(x)$ for some target network with an output $x$, the function $F$ must be smoothed in the input domain. In context of the CNN based objective functions it means the next necessary recommendations:
\begin{itemize}
\item The stride of convolutional operations is equal to one.
\item No max-pooling operations which worsen the situation with gradient artifacts. It's better to replace it with average pooling.
\item No deconvolution. The better choice is to replace single deconvolution operation by sequence of the nearest neighbor interpolation following by convolution 1x1 with stride 1.
\item Use the nonlinearity with continuous derivatives like the ELU \cite{Clevert:2015} activation function.
\end{itemize}

In our case not only the discriminator should satisfy the specified above conditions but the encoder and the attribute classifier too. As a result, we are limited in choice of the state of the art architectures to carry out the feature extraction with better quality. But as it shows in  practice the fundamental problems like GANs convergence and their diversity are more influential than the attendant problems.

Finally, the architectures of the encoder and generator implemented in this paper can be found in Tables \ref{tab:net_a} and \ref{tab:net_b}. Note, the encoder has a normalizer at the end to carry out a normalization of each pair of values according to Eq. \ref{eq:4} limitations. The generator has one additional enhancement in relation to the default architecture: according to the results in \cite{Tan:2017} we use one more upscaling step with the next average pooling operator to force the network taking into account the opinions of neighbors pixels too.

\subsection{Reconstruction loss}

The key moment from our point of view is selection of the reconstruction loss $l(;)$. Following the conclusions in \cite{Vertolli:2017} about this objective we have chosen two terms loss: $L_1$ distance for each channel in the RGB color space and gradient magnitude of Y channel in the YIQ color space:
\begin{equation} \label{eq:9}
l(x; y) = \sum_{c \in \{R,G,B\}}2|p_x^c - p_y^c| + |m_x - m_y|,
\end{equation}
where $p_x^c$ - pixel value of $c$ channel of image $x$ and $m_x$ - the gradient magnitude of image $x$.

The next problem during training is over-fitting on most frequent templates which in case of face synthesis problem is more actual due to common similarity of faces.

In general the best solution is to focus on the most difficult spatial areas of the reconstructed image i.e. areas with the high reconstruction error $l(;)$. In our opinion, the most suitable solution was proposed in \cite{Bulo:2017} by introducing the Loss Max-Pooling operation $l_{mp}(;)$. This operation allows to filter the pixels with low value of errors and being focused on each hard areas.

Regarding implementation of the above declared method we have modified the paradigm of pixel set selection - originally the loss max-pooling acts on each image from the batch independently but we have merged pixel-losses from all images in one set. It means that whole images can be skipped if current sample is already well known during the optimization step. Our experiments showed that it enhances the situation with possible over-fitting in case of training on small dataset.

\subsection{Generator architecture}

Having the wide choice of possible GAN frameworks to use in our network we have focused on BEGAN. The main reason to use BEGAN is it's the paradigm to measure the reconstruction loss as a GAN objective instead of commonly used binary answer i.e. true or false for sources of the input distribution.

The BEGAN based discriminator uses the same idea of auto-encoders which is much discussed above. One more change to the original BEGAN discriminator which we have made is its extending for the conditional inference. This was achieved by simple concatenation of attribute vector to channels after unfolding an output of a dense layer into a spatial representation.

The negative side of such decision is low sensitiveness of final solution to the input attributes. We believe that it's a property of BEGAN framework with the proposed $l_{mp}(;)$ loss and without the degradation in the image performance the improvements are unlikely.

In this evidence we are forced to incorporate the additional subnetwork named \textit{attribute classifier} aiming to control the conformity of the generated images to the initial attributes.

The discriminator $D$ architecture is a concatenation of an encoder and generator architectures (see Table \ref{tab:net_b}) but without normalization layer in the encoder and the last up-sampling block in a generator.

\begin{figure}[t]
\centering
\includegraphics[width=0.6\textwidth]{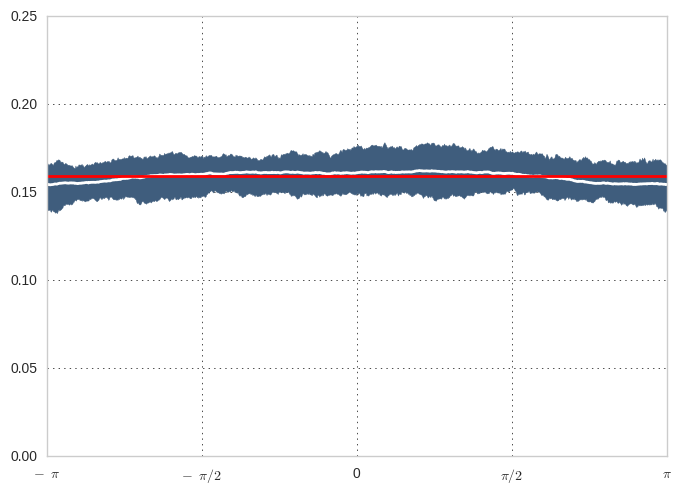}
\caption{Density of the latent space variables in case of common uniform distribution. The red color curve is the theoretical distribution and white one with blue bars - the learned distribution.}
\label{fig:out_dist}
\end{figure}

\subsection{Attribute classifier}

In this work we deal with three types of face attributes: age, gender and ethnicity. The last two are the categorical variables and can be modeled by the simple \textit{softmax} function. Regarding the first one it's a continuous variable. In literature it's proposed to tackle with it either as a regression task or classification over the quantized values. In practice the last one works better however the target objective should take into account the neighborhood property - the closer the prediction to the true label, the lower the value of loss function should be and vise versa.

In this work we use combined loss for the quantized values of an age attribute variable. It consists of three terms: simple \textit{softmax} for the classification objective, the \textit{softargmax} \cite{Xavier:2013} for the regression and the \textit{locality loss} (see Eq. \ref{eq:14}, where $y_i$ true values of age) to force the prediction be local near the true age value.
\begin{equation} \label{eq:14}
\textit{locality}(x) = \sum_{i} \frac{e^{\beta x_i}}{\sum_{j} e^{\beta x_j}}|i - y_i|.
\end{equation}

One more challenge we have faced is combining losses from different networks heads: age, gender, ethnicity attributes predictors. The simple solution like re-weighting losses manually is not working well properly because of the possibility to dominate one objective over the rest during training. Reasonable solution for this problem is to use adaptive scheme of loss mixing. We have chosen two term solution: proposed in \cite{Lin:2017} \textit{focal loss} to tackle with class imbalance problem and the adaptive multi-target loss with learnable weights imbued by the idea from \cite{Kendall:2017}. Assume below $\{v_1^t, ..., v_n^t\}$ is a set of nonnegative values from the time step $t$. We can interpret each independent value as \textit{time serie} $(v_i^1, ..., v_i^t)$ and the best practice to work with such data is applying exponential smoothing, which gives us a set of smoothed values $\{s_1^t, ..., s_n^t\}$. Than we re-weight each value $v_i^t$ by multiplying with regularization weight $w_i^t$ and learnable weight $\gamma_i^t$:
\begin{equation} \label{eq:16}
l_{ada}(v^t) = \sum_{i=1}^{n} \frac{(\gamma_i^t)^2 v_i^t w_i^t}{\sum_{k = 1}^{n} (\gamma_k^t)^2}, w_i^t = \frac{\sum_{k = 1}^{n} s_k^t}{n s_i^t}.
\end{equation}

The next question for us is how to incorporate the attribute classifier into the main pipeline. There are two obvious variants: to train the best attribute classifier separately from the main network and then to add it "as is" to propagate the gradients through it only or to merge it in one super-net and to optimize it in adversarial manner as GAN framework too. In case of the last one the objective on the negative phase (adversarial) of GAN optimizing will be default cross-entropy with the uniform labels for the true distribution.

We have tried with both variants and figure out that the best solution in our case is to train the attribute classifier as an individual network but with strong augmentation of the input images, dropouts for the last two fully-connected layers and additional penalty to the input sensitivity (Frobenius norm of the Jacobian of the nonlinear mapping) as in the contractive auto-encoder (originally \cite{Rifai:2011}).
\par The attribute classifier $A$ architecture is demonstrated in Table \ref{tab:net_c}. The network consists in the shared backbone (in the table it's everything before the last block) and the class specific heads (last block in each table, where $n_{classes}$ is number classes in the current task e.g. for the gender classification task $n_{classes} = 2$).

\subsection{Optimization}

All experiments have been performed in the TensorFlow framework \cite{Abadi:2015}. Both the attribute classifier and the main networks have been optimized using batch size 32 with Adam optimizer and exponentially decayed learning rate starting with $10^{-4}$.

Regarding the whole network training we use the same method as in BEGAN paper \cite{Berthelot:2017} and update the parameters $\theta_E$, $\theta_G$ and $\theta_D$ independently on the same step and no alternating for $G$ and $D$ was used.

Note that the end-to-end training with the proposed losses is more sophisticated task than default GAN training even with its convergence problems. This fact can be explained by problem of balance between target GAN training and the convergence of the additional objectives.

\section{Experimented Result}

\subsection{Data}

To evaluate the proposed method we have chosen a UTKFace dataset \cite{Song:2017} because it contains challengeable labels like categorical age, ethnicity and continues age in addition to the large scale face data . This dataset includes more than 20k images in the wild. The ground truth of an annotation is estimated by the DEX algorithm \cite{Rothe:2015} and double checked by humans as it's reported by the authors.

To demonstrate the power of the proposed solution we decided to carry out experiments on the aligned and cropped images of faces thereby complicating the task in terms of variety of faces because there is no impact of the secondary attributes like hairstyle and dress on the face diversity.

\subsection{Face synthesis}

The first thing we should discuss is properties of the learned distribution of the latent vector $z$. It was demonstrated above that the default choice of a space for the latent vector either restricts the support size like for the uniform distribution or produces significantly different distribution (like for the normal distribution), making sampling impossible.

However after changing the latent space to the unit complex number and applying the introduced losses the final distribution is well-fitted to the theoretical as it's shown on the Fig. \ref{fig:out_dist}. In other words, after handling this we can generate the images from the same distribution as it was trained on and no other transformation is required. The summary of using other latent spaces is shown in the Table \ref{tab:comp_z}.

\begin{table}
\caption{Results of training with different latent spaces.}
\centering
\begin{tabular}{ r | l }
 \textbf{Latent space} & \textbf{Result} \\
 \hline
 Gaussian $N(0; 1)$ & Not converged \\
 Uniform $U[-1; 1]$ & Capacity degradation \\
 VAE \cite{Kingma:2013} & Initially low capacity \\
 \hline
 \textbf{Unit complex numbers} & \textbf{Maximal diversity} \\
\end{tabular}
\label{tab:comp_z}
\end{table}

The next challenge is to make the model be more sensitive to the attribute changing task. The problem is complicated by the fact that synthesized faces must fit the same personality in addition to the general requirement to be photo-realistic. We have performed several experiments to show the robustness of our solution for the specified above problem: variation of the gender (Fig. \ref{fig:var_gender_ethno}, left), ethnicity (Fig. \ref{fig:var_gender_ethno}, right) and age (Fig. \ref{fig:var_age}) attributes.

Finally, Fig. \ref{fig:gen_faces} shows examples of generated images with uniform attributes. As you can see even normalized face images (crop and align transforms) look well enough providing a good trade-off between photo-realistic and diversity properties.

\subsection{Diversity test}

As it was mentioned before, the most important question addressed to GANs is the diversity of the generated samples or in other words the size of the \textit{support} of the discrete distribution. The simple but powerful framework for the support size estimation is based on the \textit{birthday paradox}. It says if the discrete distribution support size is $N$ then set of $\sqrt{N}$ samples should have duplicates.

\begin{table}
\caption{The "birthday paradox" test results of different methods. Proposed solution demonstrates the diversity compatible with ALI method but on the significantly higher resolution and relatively small train dataset.}
\centering
\begin{tabular}{ c | c | c | c }
 \textbf{Method} & \textbf{Resolution} & \textbf{Train size} & \textbf{Diversity} \\
 \hline
 DCGAN \cite{Radford:2015} & 32x32 & 200k & 0.8x \\
 MIX+DCGAN \cite{Arora:2017b} & 32x32 & 200k & 0.8x \\
 ALI \cite{Dumoulin:2016} & 32x32 & 200k & 5x \\
 \hline
 \textbf{Our} & \textbf{128x128} & \textbf{20k} & \textbf{3x} \\
\end{tabular}
\label{tab:comp_m}
\end{table}

To cope with it in the context of GANs diversity the \textit{birthday paradox test} has been proposed in \cite{Arora:2017}. To measure the difference between the generated samples in an automatic manner the FaceNet \cite{Schroff:2015} is used. It outputs the embedding for each input face image and decides whether the faces are same or not (using the $L_2$ distance between them). Finally, the test includes the next simple steps:
\begin{enumerate}
\item Sample the set of $N$ images using the generator $G$.
\item Select top $k$ (relatively small number to check by hand) most similar pairs of faces using FaceNet.
\item Check duplicates in top $k$ pairs by hand.
\item If no duplicates are found then increase $N$ and repeat. Otherwise stop and the support size is $N^2$.
\end{enumerate}

Regarding the proposed in this paper solution we have found that with the probability higher than $50\%$ the batch of 250 images in resolution 128x128 contains duplicates. It means that the support size is about $60k$ samples and it's three times more than the training set size. The results of other methods (according to the work \cite{Arora:2017}) are presented in the Table \ref{tab:comp_m}.

\section{Conclusion}

In this work we highlighted the problem of generating photo-realistic faces depending on face attributes. We demonstrated the efficiency of the auto-encoder based solution which resolved the diversity problem of generated by the GAN framework images and focused the attention on the mode collapse problem. To cope with it we introduced the novel space for the latent vector and the auxiliary loss for efficient fitting of the target distribution. Finally, we empirically evidenced that our method can generate the attribute guided face images with the desired trade-off between the quality and diversity showing the support size of the high-resolution 128x128 images 3 times higher than the size of original dataset. We believe that this work introduces one more helpful tool to work with auto-encoders in general case and with GANs in particular.

\bibliographystyle{splncs}
\bibliography{egbib}

\begin{appendix}
\counterwithin{figure}{section}
\counterwithin{table}{section}

\section{Final images}

\begin{figure}
\centering
\includegraphics[height=4cm]{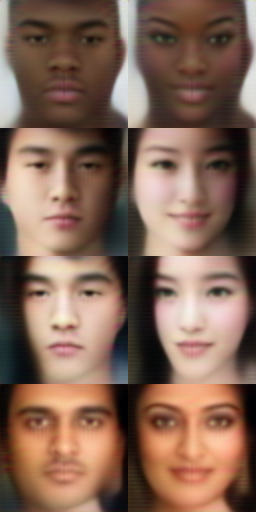}
\hspace{30pt}
\includegraphics[height=4cm]{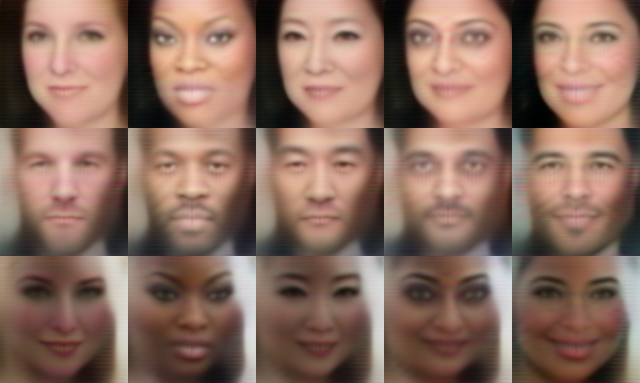}
\caption{Left: variation of the gender attribute line by line. Right: variation of the ethnicity attribute (in each line the same person)}
\label{fig:var_gender_ethno}
\end{figure}

\begin{figure}
\centering
\includegraphics[width=\textwidth]{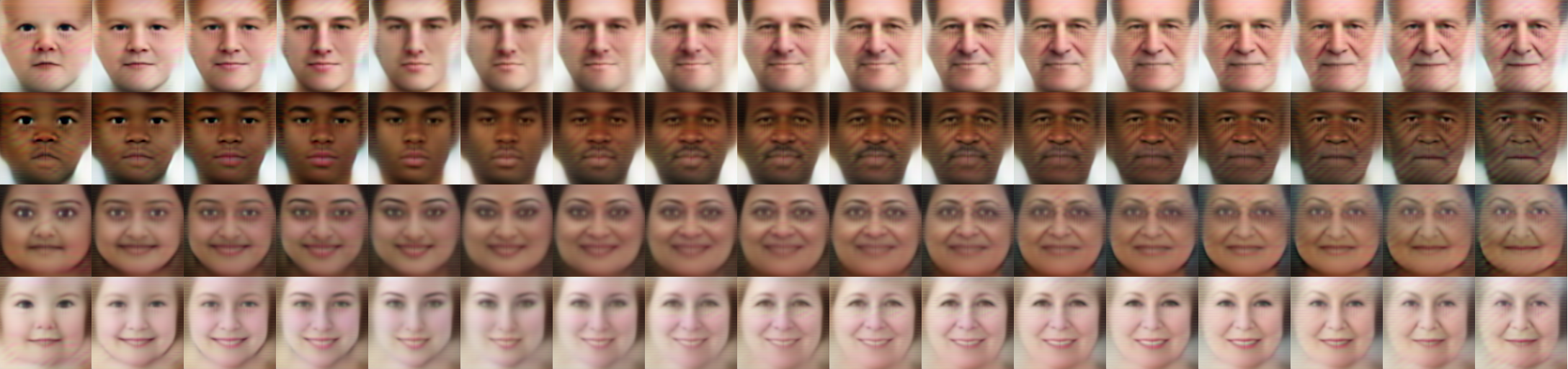}
\caption{Variation of the age attribute for the same random person.}
\label{fig:var_age}
\end{figure}

\begin{figure}
\centering
\includegraphics[width=\textwidth]{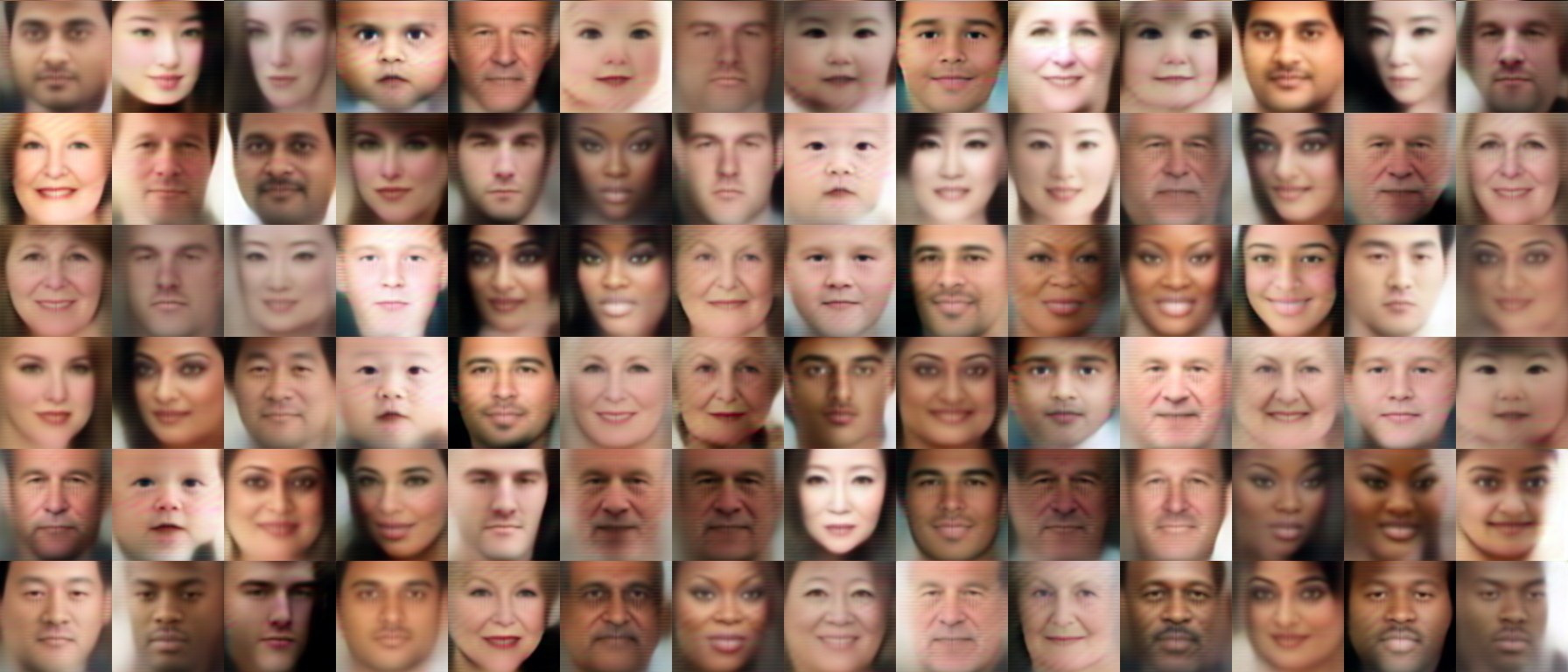}
\caption{Generated face images.}
\label{fig:gen_faces}
\end{figure}

\clearpage

\section{Network architectures}

Layer name shortcuts: \textit{conv} - convolution operator with next batch norm and ELU nonlinearity, \textit{avg pool} - average pooling operator, \textit{fc} - fully connected layer, \textit{norm} - pairwise normalizer, \textit{nn resize} - nearest neighbor resize operator.

\begin{table}

\centering
\caption{Encoder $E$ architecture}
\begin{tabular}{ l | c | c | l }
 \textbf{Layer} & \textbf{Kernel} & \textbf{Stride} & \textbf{Out HxWxC} \\ \hline
 input & - & - & 128 x 128 x 3 \\ \hline
 conv & 3x3 & 1 & 128 x 128 x 64 \\
 conv & 3x3 & 1 & 128 x 128 x 64 \\
 avg pool & 2x2 & 2 & 64 x 64 x 64 \\
 conv & 3x3 & 1 & 64 x 64 x 64 \\ \hline
 conv & 3x3 & 1 & 64 x 64 x 128 \\
 conv & 3x3 & 1 & 64 x 64 x 128 \\
 avg pool & 2x2 & 2 & 32 x 32 x 128 \\
 conv & 3x3 & 1 & 32 x 32 x 128 \\ \hline
 conv & 3x3 & 1 & 32 x 32 x 256 \\
 conv & 3x3 & 1 & 32 x 32 x 256 \\
 avg pool & 2x2 & 2 & 16 x 16 x 256 \\
 conv & 3x3 & 1 & 16 x 16 x 256 \\ \hline
 conv & 3x3 & 1 & 16 x 16 x 512 \\
 conv & 3x3 & 1 & 16 x 16 x 512 \\
 avg pool & 2x2 & 2 & 8 x 8 x 512 \\
 conv & 3x3 & 1 & 8 x 8 x 512 \\ \hline
 conv & 3x3 & 1 & 8 x 8 x 1024 \\
 conv & 3x3 & 1 & 8 x 8 x 1024 \\
 avg pool & 2x2 & 2 & 4 x 4 x 1024 \\
 conv & 3x3 & 1 & 4 x 4 x 1024 \\ \hline
 fc & - & - & 1 x 1 x 100 \\ \hline
 norm & - & - & 1 x 1 x 100
\end{tabular}
\label{tab:net_a}
\end{table}

\begin{table}
\centering
\caption{Generator $G$ architecture}
\begin{tabular}{ l | c | c | l }
 \textbf{Layer} & \textbf{Kernel} & \textbf{Stride} & \textbf{Out HxWxC} \\ \hline
 input & - & - & 1 x 1 x 100 \\ \hline
 fc & - & - & 1 x 1 x 16384 \\ \hline
 reshape & - & - & 4 x 4 x 1024 \\ \hline
 conv & 3x3 & 1 & 4 x 4 x 512 \\
 conv & 3x3 & 1 & 4 x 4 x 512 \\
 nn resize & - & - & 8 x 8 x 512 \\ \hline
 conv & 3x3 & 1 & 8 x 8 x 256 \\
 conv & 3x3 & 1 & 8 x 8 x 256 \\
 nn resize & - & - & 16 x 16 x 256 \\ \hline
 conv & 3x3 & 1 & 16 x 16 x 128 \\
 conv & 3x3 & 1 & 16 x 16 x 128 \\
 nn resize & - & - & 32 x 32 x 128 \\ \hline
 conv & 3x3 & 1 & 32 x 32 x 64 \\
 conv & 3x3 & 1 & 32 x 32 x 64 \\
 nn resize & - & - & 64 x 64 x 64 \\ \hline
 conv & 3x3 & 1 & 64 x 64 x 32 \\
 conv & 3x3 & 1 & 64 x 64 x 32 \\
 nn resize & - & - & 128 x 128 x 32 \\ \hline
 conv & 3x3 & 1 & 128 x 128 x 32 \\
 nn resize & - & - & 256 x 256 x 32 \\
 conv & 3x3 & 1 & 256 x 256 x 32 \\ \hline
 conv & 3x3 & 1 & 256 x 256 x 3 \\ \hline
 avg pool & 3x3 & 2 & 128 x 128 x 3
\end{tabular}
\label{tab:net_b}
\end{table}

\begin{table}
\centering
\caption{Attribute classifier $A$ architecture}
\begin{tabular}{ l | c | c | l }
 \textbf{Layer} & \textbf{Kernel} & \textbf{Stride} & \textbf{Out HxWxC} \\ \hline
 input & - & - & 128 x 128 x 3 \\ \hline
 conv & 3x3 & 1 & 128 x 128 x 16 \\
 avg pool & 2x2 & 2 & 64 x 64 x 16 \\ \hline
 conv & 3x3 & 1 & 64 x 64 x 16 \\
 avg pool & 2x2 & 2 & 32 x 32 x 16 \\ \hline
 conv & 3x3 & 1 & 32 x 32 x 32 \\
 avg pool & 2x2 & 2 & 16 x 16 x 32 \\ \hline
 conv & 3x3 & 1 & 16 x 16 x 48 \\
 avg pool & 2x2 & 2 & 8 x 8 x 48 \\ \hline
 conv & 3x3 & 1 & 8 x 8 x 64 \\
 avg pool & 2x2 & 2 & 4 x 4 x 64 \\ \hline
 conv & 3x3 & 1 & 4 x 4 x 96 \\
 conv & 3x3 & 1 & 4 x 4 x 128 \\ \hline
  conv & 3x3 & 1 & 4 x 4 x 192 \\
 avg pool & 4x4 & 1 & 1 x 1 x 192 \\
 fc & - & - & 1 x 1 x 192 \\
 fc & - & - & 1 x 1 x 192 \\
 fc & - & - & 1 x 1 x $n_{classes}$ \\
\end{tabular}
\label{tab:net_c}
\end{table}

\clearpage

\section{Training scheme}

\begin{figure}
\centering
\includegraphics[width=\textwidth]{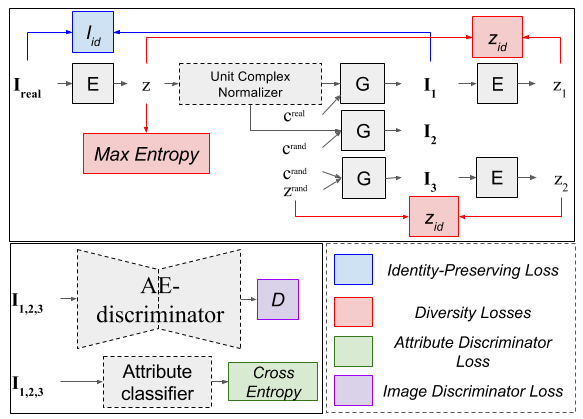}
\caption{Network training scheme. The proposed scheme for training a face generator with the auto-encoder based identity and diversity preserving losses. Rectangles with the same subnetwork names (e.g. encoder $E$) have shared parameters. On the figure $c^{real}$ and $c^{rand}$ - the real and random attribute labels correspondingly, $z^{rand}$ - the random normalized latent vector, $I_{real}$ - an input train sample, $I_{1}$, $I_{2}$, $I_{3}$ - the generated images, $I_{id}$ - the reconstruction loss in the image domain, $z_{id}$ - the reconstruction loss in the latent space domain.}
\label{fig:architecture}
\end{figure}

\end{appendix}

\end{document}